\documentclass[conference]{IEEEtran}
\IEEEoverridecommandlockouts

\usepackage{cite}
\usepackage{amsmath,amssymb,amsfonts}
\usepackage{algorithmic}
\usepackage{graphicx}
\usepackage{textcomp}
\usepackage{xcolor}
\usepackage{multirow, multicol, mathtools, lipsum}
\usepackage{colortbl}
\usepackage{xcolor}
\usepackage{float}
\usepackage{graphicx}

\def\BibTeX{{\rm B\kern-.05em{\sc i\kern-.025em b}\kern-.08em
    T\kern-.1667em\lower.7ex\hbox{E}\kern-.125emX}}
\begin{document}

\title{Continuous Adversarial Text Representation Learning for Affective Recognition

\thanks{This work was supported by the Korea Institute for Advancement of Technology (KIAT) grant funded by Ministry of Trade, Industry and Energy (MOTIE) (No. P0028821, Development of 22kW Variable Type EV Charging System for Building Up Global Chain of Intelligent/Universal Mechanical Parking System).
}
}

\author{\IEEEauthorblockN{1\textsuperscript{st} Seungah Son}
\IEEEauthorblockA{\textit{CCS Graduate School of Mobility} \\
\textit{Korea Advanced Institute of } \\ \textit{ Science and Technology (KAIST)}\\
Daejeon, Rep. of Korea, 34141 \\
seungahson@kaist.ac.kr}  
\and

\IEEEauthorblockN{2\textsuperscript{nd} Andres Saurez}
\IEEEauthorblockA{\textit{Robotics Program} \\
\textit{Korea Advanced Institute of } \\ \textit{ Science and Technology (KAIST)}\\
Daejeon, Rep. of Korea, 34141 \\
asaurez@kaist.ac.kr}  
\and

\IEEEauthorblockN{3\textsuperscript{rd} Dongsoo Har}
\IEEEauthorblockA{\textit{CCS Graduate School of Mobility} \\
\textit{Korea Advanced Institute of } \\ \textit{ Science and Technology (KAIST)}\\
Daejeon, Rep. of Korea, 34141 \\
dshar@kaist.ac.kr}  
}

\maketitle

\begin{abstract}
While pre-trained language models excel at semantic understanding, they often struggle to capture nuanced affective information critical for affective recognition tasks. To address these limitations, we propose a novel framework for enhancing emotion-aware embeddings in transformer-based models. Our approach introduces a continuous valence-arousal labeling system to guide contrastive learning, which captures subtle and multidimensional emotional nuances more effectively. Furthermore, we employ a dynamic token perturbation mechanism, using gradient-based saliency to focus on sentiment-relevant tokens, improving model sensitivity to emotional cues. The experimental results demonstrate that the proposed framework outperforms existing methods, achieving up to 15.5\% improvement in the emotion classification benchmark, highlighting the importance of employing continuous labels. This improvement demonstrates that the proposed framework is effective in affective representation learning and enables precise and contextually relevant emotional understanding.

\end{abstract}


\section{Introduction}
Affective recognition has gained significant attention in the field of natural language processing (NLP), with a wide range of applications, including human-computer interaction, recommendation systems, and management information systems \cite{Cambria2023, Afzal2023}. To date, as a subfield of affective recognition, emotion recognition has been investigated with flat or hierachical sensor networks \cite{ip1,sur1}. There is a growing interest in making human-device interactions more natural and refined through AI-driven technologies. This emphasis aims to enhance the quality of interaction between machines and humans, making it more seamless and sophisticated. In light of these advances, a comprehensive understanding of the entire range of human emotional responses in digital communications has become increasingly essential. \cite{EmoRec}

Beginning with the application of deep learning methodologies \cite{KimReview}, pre-trained language models (PLMs), particularly those based on transformer architectures like BERT \cite{BERT} and GPT \cite{GPT}, have transformed the field of NLP by effectively extracting semantic information and producing contextualized text representations. Many sentiment analysis approaches have adopted PLMs as a base model, treating sentiment-related applications as downstream tasks. However, while PLMs excel at semantic understanding, they often struggle to capture nuanced affective information critical to sentiment analysis tasks without careful fine-tuning \cite{pretraining}. 

To address this problem, recent studies have focused on the development of sentiment-aware pre-trained language models by modifying existing methodologies employed for semantic similarity and incorporating sentiment information.
Although these enhancements have shown promising results in downstream sentiment tasks, existing sentiment analysis methods still face challenges in capturing nuanced emotional distinctions due to several factors.
Their reliance on binary or ternary polarity labels oversimplifies the complex, multidimensional nature of human emotions, risking dimensional collapse and reducing generalization across contexts. Furthermore, dependence on static lexicons introduces biases and hampers adaptability to domain, cultural, and linguistic variations. In addition, narrow masking strategies focus on small subsets of input tokens, failing to account for broader emotional patterns within global text structures. These limitations collectively restrict the effectiveness of current sentiment analysis models.

\color{black}
In this paper, we propose continuous adversarial representation learning (CARL), which leverages transformer-based text encoder training to enhance emotion-aware embeddings.
The main contributions of our work can be summarized as follows:

\begin{itemize}
\item We propose a novel training framework for affective recognition using sentence-level momentum continuous label contrastive learning and token-level dynamic adversarial attack detection guided by gradients.

\item Our approach boosts models' affective embedding capabilities, contributing to more robust and nuanced emotion-aware text representations.

\item We achieve superior performance compared to baseline methods across three key emotion recognition tasks.

\item Furthermore, our method yields higher quality in the emotional representation space, tested with the balance between alignment and uniformity of the emotions.
\end{itemize}

\section{Related Works}
Text representation learning aims to extract meaningful information from textual data, employing metric learning techniques that embed objects into spaces where semantic relationships are preserved. These embeddings are critical for downstream tasks, as they facilitate the model's ability to generalize across various applications. Text representation learning can generally be divided into two main categories: intra-sample and inter-sample predictions.

Intra-sample prediction focuses on creating self-supervised tasks within individual samples. This approach has become popular with the introduction of BERT which uses masked language modeling (MLM) as its primary task. In MLM, tokens are masked, and the model learns to predict them using contextual information from the surrounding tokens. The BERT post-training (BERT-PT) \cite{BERT-PT} fine-tunes the model with sentiment-specific objectives using large-scale review datasets and tailors the learned representations to capture sentiment nuances. The SentiBERT \cite{SentiBERT} extends these ideas by introducing a parse tree on top of the attention mechanisms and allows the BERT to incorporate syntactic structures into its sentiment representations. This integration facilitates a deeper understanding of the sentiment relationships embedded in complex sentence structures. The SentiLARE \cite{SentiLARE} leverages SentiWordNet lexical semantics and embeds linguistic features such as sentiment scores at the word level directly into the model to enhance its ability to understand words with sentiments. The SentiX \cite{SentiX} incorporates masked emotional indicators such as emojis and sentiment-bearing words for prioritizing emotionally significant tokens in training.

Inter-sample prediction emphasizes relationships between different samples, with contrastive learning emerging as a pivotal mechanism for this purpose. Since the introduction of SimCSE \cite{Gao2021}, which demonstrated the efficacy of contrastive learning in sentence embedding tasks with minimal data augmentation, this paradigm has gained significant traction in natural language processing. Building on this foundation, SCAPT\cite{Li2021} adopts supervised contrastive learning to align representations of sentiment expressions with the same polarity, effectively capturing both implicit and explicit sentiment orientations. Similarly, SentiCSE \cite{SentiCSE} extends the contrastive learning framework by integrating masked token polarity prediction, enabling fine-grained alignment of sentiment information at both token and sentence levels. 

\section{Proposed Method}
\begin{figure*}[htbp!]
    \centering
    \includegraphics[width=0.8\textwidth]{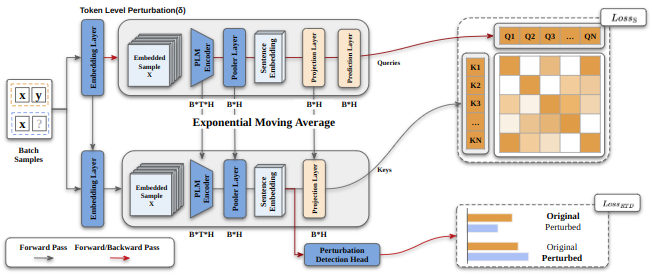}
    \caption[Full Proposed Method]{Structure of proposed framework}
    \label{fig:FullProposed}
    \vspace{-10pt}
\end{figure*}

To overcome the limitations of the existing methods, the proposed framework integrates two pre-training tasks at the sentence and token levels, as illustrated in Fig. \ref{fig:FullProposed}. For sentence-level inter-sample training, momentum continuous contrastive learning (MCCL) is used, and for token-level intra-sample training, gradient-based perturbed token detection (PTD) is utilized. 

\subsection{Momentum Contrastive Learning Structure}
Previous sentiment representation studies often employed the SimCSE framework, which relies on direct encoder updates. However, such updates can introduce instability in embedding updates and high sensitivity to batch size, hindering the learning of robust representations.

We propose adopting a momentum update strategy inspired by BYOL \cite{grill2020bootstrap}. The method employes two neural networks: an online network and a target network, both of which share the same architecture but have distinct roles. The online network learns representations through standard back propagation, while the target network provides stable representations to guide the online network's learning. The target network's parameters, denoted as $\theta_t$, are not updated directly by gradients. Instead, they are updated using an exponential moving average of the online network's parameters $\theta_o$ given as

\begin{equation}
\theta_t \leftarrow m \theta_t + (1-m) \theta_o
\end{equation}
where $m \in [0,1]$ is the momentum coefficient. To ensure dynamic adaptation, $m$ is updated during training using a cosine annealing scheduler, for a given training step $k$ out of a total of $K$ steps, so the updated $m$, $m_{\text{new}}$, is obtained as

\begin{equation}
m_{\text{new}} = 1 - (1 - m_{\text{initial}}) \cdot \frac{\cos(\pi \cdot k / K) + 1}{2}
\end{equation}

The embedding and the encoding layer in the online and target networks are identical to those of the base model. However, the activation function in the pooler layer is removed, considering additional layers added afterward. The projection layer is responsible for mapping the outputs of both the online and target branches into a shared latent space suitable for contrastive learning. The prediction layer further refines the outputs by aligning the features generated by the online network with those of the target network, which is critical for preventing model collapse. 

\subsection{Continuous Label Contrastive Learning}

We propose a continuous label contrastive learning framework grounded in Russell's circumplex model of affect \cite{Russell2022}. This model represents emotions within a two-dimensional space defined by valence and arousal, where valence characterizes the pleasure-displeasure axis and arousal denotes the activation-deactivation spectrum, indicating emotional intensity. By leveraging these multidimensional labels, our approach captures nuanced emotional variations and relationships that binary classification methods cannot. This extends the capability of conventional contrastive learning techniques, enabling a more refined understanding of emotional content.

In our framework, we utilize the in-batch similarity matrix commonly employed in self-supervised contrastive learning. For a given pair of embeddings, $z_i$ and $z_j$, obtained from the online and target networks, their similarity is computed using cosine similarity given as

\begin{equation} I_{\text{sim},ij} = \frac{ z_{i} \cdot z_{j}}{\sqrt{z_{i}^2} \cdot \sqrt{z_{j}^2}}. \end{equation}

Similarly, the label similarity matrix using the valence-arousal labels is calculated using cosine similarity defined as:

\begin{equation} I_{va,ij} = \frac{v_i v_j + a_i a_j}{\sqrt{v_i^2 + a_i^2} \cdot \sqrt{v_j^2 + a_j^2}}. \end{equation}

where $v$ and $a$ represent valence and arousal, respectively, and $i$ and $j$ denote individual samples in the dataset.

To ensure compatibility, both matrices are normalized into probability distributions using the softmax function given by:

\begin{equation} P_{x,ij} = \frac{\exp(I_{x,ij})}{\sum_k \exp(I_{x,ik})}  \end{equation}
where $x \in {\text{sim}, \text{va}}$.

We adopt cross-entropy to achieve alignment of relative proportions between distributions. This approach offers a more targeted mechanism for aligning distributions, prioritizing their structural consistency or shape over mere numerical discrepancies. Therefore, the sentence-level loss is formulated as the symmetric cross entropy between the label similarity matrix and the embedding similarity matrix as follows:

\begin{align}
\mathcal{L}_{\text{MCCL}} 
&= -\frac{1}{N} \sum_{j=1}^{N} \left[ \sum_{i=1}^{N} p_{\text{sim,i,j}} \log p_{\text{va,i,j}} + \sum_{i=1}^{N} p_{\text{va,i,j}} \log p_{\text{sim,i,j}} \right] \notag \\
&= -\frac{1}{N} \sum_{j=1}^{N} \left[\text{CE}(p_{\text{sim,j}}, p_{\text{va,j}}) + \text{CE}(p_{\text{va,j}}, p_{\text{sim,j}}) \right]
\end{align}

where $\text{CE}$ denotes the cross entropy loss. This symmetric formulation ensures alignment between embedding similarity and the underlying valence-arousal labels, promoting a robust representation of nuanced emotional content.
  
\subsection{Gradient-Based Perturbed Token Detection}

Instead of relying on traditional random masking or predefined lexicon masking, which limits the efficiency of training to only a portion of the sample, this work proposes using saliency derived from gradients to identify tokens with the highest sentiment relevance dynamically. Inspired by the replaced token detection method employed in ELECTRA \cite{ELECTRA}, this framework uses gradient-based perturbations instead of requiring a generator, thus avoiding the complexities of generator-discriminator structures. During the forward pass of the online branch, gradients are captured for each token $e \in \mathbb{R}^d$. The importance score for each token is then determined by the magnitude of its gradient ($g_t$), given by

\begin{equation}
    g_t = \left\|\frac{\partial \mathcal{L}}{\partial e_t}\right\|_2, \quad t \in \{1, \dots, T\},
\end{equation}
where $T$ is the sequence length and $\mathcal{L}$ is the loss function. 

Then, we sort and select the top-$k$ of tokens with the highest gradient norms $s_i$ to form the set of important tokens $\mathcal{S}$ and create a mask $m_t$. Here, we create the adversarial perturbation samples using the projected gradient descent (PGD) \cite{PGD} per token in the embedding layer, which is a white box attack that causes the model to make incorrect predictions. The embedding of $e_t$ is given by

\begin{equation}
    e_t^{(i+1)} = \Pi_{\mathcal{B}(e_t, \epsilon)} \left(e_t^{(i)} + \alpha \text{sign}(\frac{\nabla_{e_t} \mathcal{L}}{\|\nabla_{e_t} \mathcal{L}\|_2}) \right),
\end{equation}
where $\Pi_{\mathcal{B}(e_t, \epsilon)}$ projects the perturbed embedding back into the $\epsilon$-ball around the original embedding $e_t$, and $\alpha$ is the step size. The subtle attacks result in alterations to the token values, leading to a loss of their original semantic integrity. To prevent these perturbations from surpassing a predefined threshold, they are constrained with Frobenius norm projection.

The final embeddings can be expressed as
\begin{equation}
    \tilde{e}_t =
    \begin{cases} 
      e_t + \Delta e_t & \text{if } t \in \mathcal{S}, \\
      e_t & \text{otherwise}.
    \end{cases}
\end{equation}
where $\tilde{e}_t$ represents the total perturbation matrix. Finally, the replaced token detection layer is trained to determine whether a token has been perturbed, employing a final sigmoid activation function in this process. To address the issue of class imbalance, focal binary loss is utilized

\begin{equation}
    \mathcal{L}_{\text{PTD}} = -(1 - p_t)^\gamma \log(p_t),
\end{equation}
where $p_t$ denotes the layer's prediction of the token's susceptibility to attack, and $\gamma$ is the parameter emphasizing difficult-to-classify classes.


\subsection{Total Loss Function}

In summary, the overall loss combines the momentum contrastive learning and perturbation token detection losses:
\begin{equation}
    \mathcal{L} = \lambda_1 \mathcal{L}_{\text{MCCL}} + \lambda_2 \mathcal{L}_{\text{PTD}},
\end{equation}
where \(\lambda_1\) and \(\lambda_2\) are weighting factors that control the relative importance of each loss component.

\section{Experiments}
\subsection{Implementation Details}
We implemented the proposed method using PyTorch and trained it on an NVIDIA RTX A6000 GPU for 2 epochs with a batch size of 128. The evaluation was conducted using two transformer-based architectures: BERT-base-uncased and RoBERTa-base-uncased. The models were optimized with the AdamW optimizer, with a cosine annealing learning rate schedule with warm restarts and a 10\% warmup period, initialized with a learning rate of \(2 \times 10^{-5}\).
We use the first token (CLS) pooling for the output of the model, with the temperature parameter set to 0.05. A maximum sequence length of 128 tokens is maintained, with the momentum coefficient of 0.9996. For saliency token selection, we use the top 10\% of each sentence, and the PGD’s epsilon parameter was set to 5e-9 with the alpha of 5.
The $\lambda_1$, $\lambda_2$ values are 0.8, 0.2. For every 10 steps, we evaluate the model's performance on transfer tasks to obtain the best checkpoint.

\subsection{Training Dataset}
The training data is comprised of a diverse set of corpora annotated with valence and arousal labels, capturing emotional expressions across various domains. Specifically, we utilized EmoBank \cite{EmoBank}, a dataset aggregating texts from multiple internet sources with fine-grained emotional annotations; IEMOCAP \cite{IEMOCAP}, a corpus of dyadic interactions annotated for emotion categories and dimensions; the Facebook Posts dataset \cite{FacebookPosts}, containing social media texts reflecting everyday emotional language; and EmoTales \cite{EmoTales}, a collection of folk tale sentences with emotional labels.
In total, our training set encompassed 24,392 texts with an average length of 20.63 tokens, capturing a wide range of emotional expressions across different domains. 
While the EmoBank and IEMOCAP datasets are annotated within the range of 1 to 5, and the Facebook Posts and EmoTales datasets use a scale from 1 to 9. To address the inconsistency in label ranges across datasets, min-max normalization was applied to standardize all annotations to a common scale

\begin{equation}
x_i = 2\frac{x_i - \min(x)}{\max(x) - \min(x)} - 1
\end{equation}
where \( x_i \) represents the normalized value, and the resulting range of \( x_i \) is \([-1, 1]\).

\subsection{Downstream Tasks}
Model evaluation was conducted using four comprehensive affective recognition tasks, each aimed at assessing distinct aspects of the model’s capabilities.

\paragraph{Valence and Arousal Alignment} The model's capacity to capture continuous valence and arousal labels was evaluated using linear regression models trained on sentence embeddings. Predictive accuracy was assessed via mean absolute error (MAE), while Pearson (\(r\)) and Spearman (\(\rho\)) correlations were calculated to quantify linear and monotonic relationships between predicted and ground truth values.

\paragraph{Polarity Classification}: To assess the performance of the proposed model, we applied linear probing on standard sentiment benchmarks from SentEval toolkit \cite{SentEval}, including MR, CR, SST5, and IMDB. 
These datasets encompass varying text lengths and sentiment granularity, ranging from binary classification to granular sentiment classification.

\paragraph{Emotion Classification} The ability of the model to capture emotional nuances was evaluated using the GoEmotions \cite{GoEmotions} dataset, which consists of Reddit comments annotated for 28 emotion categories. In this study, six emotions were selected based on Ekman's theory of basic emotions to facilitate a more streamlined analysis. Performance was assessed using accuracy, precision, recall, and F1-score.

\paragraph{Embedding Quality} The quality of the learned embeddings was evaluated by analyzing their alignment and uniformity \cite{wang2020understanding}. Unlike the traditional use of the semantic textual similarity benchmark, we used the GoEmotions dataset to construct positive and negative pairs.

\subsection{Experimental Results}
The proposed models demonstrated superior performance in four tasks over baseline models.

The CARL model demonstrated significant improvements in valence and arousal prediction, as shown in Table~\ref{tab:results}. 
For the BERT-based models, CARL achieved a Pearson correlation (r) of 0.721 and a Spearman correlation ($\rho$) of 0.714 for valence, representing an improvement of 7.1\% and 7.4\% over the baseline BERT model, respectively. For arousal, CARL attained r=0.629 and $\rho$=0.684, yielding improvements of 8.1\% and 6.9\%, respectively. In terms of mean absolute error (MAE), CARL reduced the error to 0.087 for valence and 0.101 for arousal, a reduction of 11.2\% and 4.7\% compared to baseline BERT, achieving the lowest MAE among BERT-based models.
For RoBERTa-based models, CARL outperformed both the baseline and sentiment-focused models, achieving r=0.741 and $\rho$=0.738 for valence, an improvement of 9.4\% and 9.8\% over baseline RoBERTa, respectively. For arousal, CARL reached r = 0.634 and $\rho$ = 0.699, representing a 2. 4\% and 3. 1\% improvement over the baseline RoBERTa. The MAE for CARL was reduced to 0.083 for valence and 0.097 for arousal, marking a 15.3\% and 5.8\% improvement compared to baseline RoBERTa.
These results indicate that both models effectively capture fine-grained affective states in text, with the RoBERTa-based model having a slight edge in performance.

\begin{table}[H] 
\caption{Performance Comparison on Valence and Arousal Alignment}
\begin{center}
\resizebox{\columnwidth}{!}{%
\begin{tabular}{|c|c|c|c|c|c|c|c|}
\hline
\multirow{2}{*}{\textbf{Base Model}} & \multirow{2}{*}{\textbf{Model}} & \multicolumn{3}{c}{\textbf{Valence}} & \multicolumn{3}{|c|}{\textbf{Arousal}} \\
\cline{3-8}
 &  & \textbf{$r$} & \textbf{$\rho$} & \textbf{MAE} & \textbf{$r$} & \textbf{$\rho$} & \textbf{MAE} \\
\hline
\multirow{5}{*}{BERT} & BERT & 0.673 & 0.665 & 0.098 & 0.582 & 0.640 & 0.106 \\
 & BERT-PT & 0.643 & 0.640 & 0.103 & 0.569 & 0.622 & 0.108 \\
 & SentiBERT & 0.675 & 0.670 & 0.098 & 0.584 & 0.635 & 0.107 \\
 & SentiX & 0.667 & 0.656 & 0.120 & 0.572 & 0.614 & 0.110 \\ 
 & \textbf{CARL} & \textbf{0.721} & \textbf{0.714} & \textbf{0.087} & \textbf{0.629} & \textbf{0.684} & \textbf{0.101} \\
\hline
\multirow{4}{*}{RoBERTa} & RoBERTa & 0.677 & 0.672 & 0.098 & 0.619 & 0.678 & 0.103 \\
 & SentiLARE & 0.697 & 0.692 & 0.095 & 0.589 & 0.665 & 0.106 \\
 & SentiCSE & 0.680 & 0.670 & 0.090 & 0.575 & 0.626 & 0.170 \\ 
 & \textbf{CARL} & \textbf{0.741} & \textbf{0.738} & \textbf{0.083} & \textbf{0.634} & \textbf{0.699} & \textbf{0.097} \\
\hline
\end{tabular}%
}
\label{tab:results}
\end{center}
\end{table}

In transfer task evaluation, the BERT-based CARL showed consistent improvements across all tasks, with particularly notable gains in fine-grained sentiment analysis. As shown in Table~\ref{tab:senteval}, the model achieved a 12.9\% improvement over base BERT on the SST5 task and a 9.2\% gain on IMDB classification. 
The RoBERTa-based CARL demonstrated even stronger transfer capabilities, showing a 10.2\% improvement over base RoBERTa on SST5 and a 2.7\% gain on IMDB. In table~\ref{tab:senteval}, cells marked with an asterisk (*) denote that the model utilized the dataset for training. 
Apart from the results annotated with an asterisk, our model outperformed others on average.
Overall, on average performance, CARL achieved a high average performance improvement of 6.5\% and 6.2\% over BERT and RoBERTa, respectively.
The balanced performance across these diverse tasks highlights the robustness and generalizability of the proposed framework. 


\begin{table}[H] 
\caption{Performance Comparison on Polarity Classification}
\vspace{-5mm}
\begin{center}
\resizebox{\columnwidth}{!}{%
\begin{tabular}{|c|c|c|c|c|c|}
\hline
\multirow{2}{*}{\textbf{Base Model}} & \multirow{2}{*}{\textbf{Model}} & \multicolumn{4}{c|}{\textbf{SentEval Transfer Tasks}}\\ 
\cline{3-6}
 &  & \textbf{CR} & \textbf{MR} & \textbf{SST5} & \textbf{IMDB} \\
\hline
\multirow{5}{*}{BERT} & BERT & 87.66 & 81.93 & 44.14 & 81.18 \\
 & BERT-PT & 88.11 & 85.59 & 48.41 & 86.48 \\
 & SentiBERT & 88.48 & 86.23 & 48.59 & 82.17 \\
 & SentiX & \textbf{91.01} & 83.57 & 44.54 & 82.48 \\
 & \textbf{CARL} & 89.17 & \textbf{86.59} & \textbf{49.83} & \textbf{88.64} \\
\hline
\multirow{4}{*}{RoBERTa} & RoBERTa & 85.91 & 82.46 &  47.50 & 89.96 \\
 & SentiLARE & 90.64 & 88.12 & 51.27 & 90.16 \\
 & SentiCSE & 90.71 & \textbf{97.41*} & 52.13 & 88.43 \\
 & \textbf{CARL} & \textbf{91.42} & 88.49 & \textbf{52.34} & \textbf{92.41} \\
\hline
\end{tabular}%
}
\label{tab:senteval}
\end{center}
\end{table}

\begin{figure}[t] 
    \centering
    \includegraphics[width=0.48\textwidth]{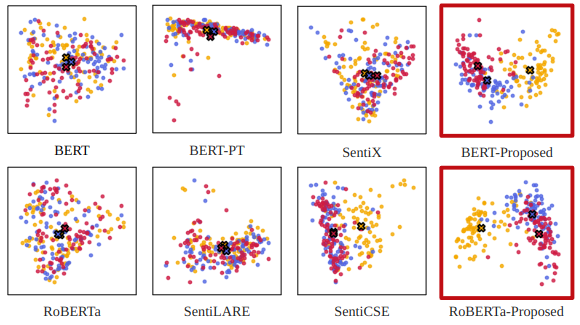}
    \caption{PCA visualization of text embeddings across models}
    \label{fig:PCAVis}
\vspace{-15pt}
\end{figure}

In emotion classification tasks, as presented in Table~\ref{tab:emotion-classification}, The BERT-based CARL significantly outperformed all BERT-variants, demonstrating a 12.9\% accuracy improvement over base BERT and a 3.6\% gain over the previous best SentiBERT. The model showed an 8.9\% improvement in the F1-score compared to base BERT and maintained consistent improvements across all metrics, with particularly strong gains in precision.
The RoBERTa-based CARL set new benchmarks in emotion classification with a 15.5\% accuracy improvement over base RoBERTa and a 3.5\% gain over the SentiCSE. The model achieved an impressive 9.3\% improvement in F1-score compared to base RoBERTa and demonstrated balanced improvements across precision and recall metrics. As visualized through PCA in Figure~\ref{fig:PCAVis}, even without extensive training, the proposed models achieve a well-separated and coherent cluster separation between emotional clusters, with the largest differences between cluster centroids. This balance underscores the representational strength of the proposed models and their adaptability to diverse downstream tasks.

\begin{table}[t] 
\caption{Performance Comparison on Emotion Classification}
\vspace{-5mm}
\begin{center}
\resizebox{\columnwidth}{!}{%
\begin{tabular}{|c|c|c|c|c|c|}
\hline
\multirow{2}{*}{\textbf{Base Model}} & \multirow{2}{*}{\textbf{Model}} & \multicolumn{4}{c|}{\textbf{Emotion Classification Metrics}} \\
\cline{3-6}
 &  & \textbf{$\mu_{\text{Acc}}$} & \textbf{$\pi_{\text{P}}$} & \textbf{$\rho_{\text{R}}$} & \textbf{$\phi_{\text{F1}}$} \\
\hline
\multirow{5}{*}{BERT} & BERT & 64.44 & 0.660 & 0.660 & 0.648 \\
 & BERT-PT & 63.17 & 0.614 & 0.612 & 0.610 \\
 & SentiBERT & 70.21 & 0.689 & 0.684 & 0.685 \\
 & SentiX & 65.71 & 0.683 & 0.682 & 0.682\\ 
 & \textbf{CARL} & \textbf{72.73} & \textbf{0.719} & \textbf{0.701} & \textbf{0.705} \\
\hline
\multirow{4}{*}{RoBERTa} & RoBERTa & 63.91 & 0.657 & 0.639 & 0.640 \\
 & SentiLARE & 71.32 & 0.716 & 0.713 & 0.713 \\
 & SentiCSE & 72.47 & 0.728 & 0.723 & 0.723 \\ 
 & \textbf{CARL} & \textbf{73.84} & \textbf{0.766} & \textbf{0.759} & \textbf{0.761} \\
\hline
\end{tabular}%
}
\label{tab:emotion-classification}
\end{center}
\vspace{-5mm}
\end{table}

\begin{figure}[H] 
    \centering
    \includegraphics[width=0.5\textwidth]{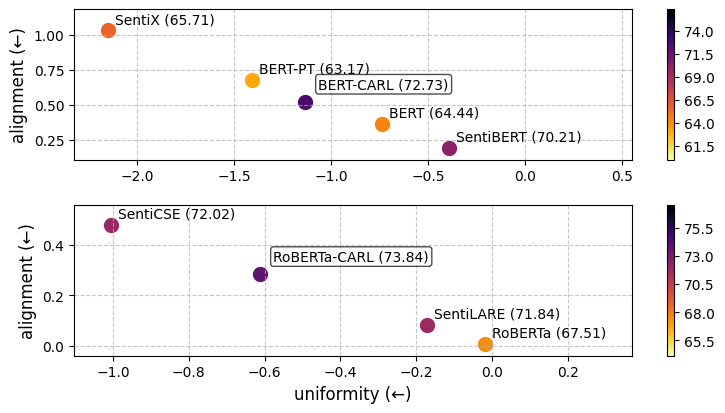}
    \caption{The alignment-uniformity plot of the models}
    \label{fig:AliUni}
    \vspace{-3mm}
\end{figure}

Figure~\ref{fig:AliUni} complements Table~\ref{tab:alignment-uniformity}, visualizing the trade-off between alignment and uniformity across all models. In the figure, the model labels are color-coded based on the classification accuracy reported in Table~\ref{tab:emotion-classification}. The proposed models demonstrate an excellent balance between these properties compared to the previous models. For example, the proposed RoBERTa model achieves an alignment score of 0.2649 and a uniformity score of -0.6063, indicating that it generates compact clusters of similar sentences while maintaining a well-dispersed embedding space. This balance underscores the representational strength of the proposed models and their adaptability to diverse downstream tasks.

\begin{table}[H]
\caption{Alignment and Uniformity Results}
\vspace{-6mm}
\begin{center}
\resizebox{\columnwidth}{!}{%
\footnotesize 
\begin{tabular}{|c|c|c|c|}
\hline
\textbf{Base Model} & \textbf{Model} & \textbf{Alignment} & \textbf{Uniformity} \\
\hline
\multirow{5}{*}{BERT} & BERT & 0.3633 & -0.7383 \\
 & BERT-PT & 0.6776 & -1.4091 \\
 & SentiBERT & 0.1907 & -0.3892 \\
 & SentiX & 1.0338 & -2.1505 \\ 
 & \textbf{CARL} & \textbf{0.5027} & \textbf{-1.1224} \\
\hline
\multirow{4}{*}{RoBERTa} & RoBERTa & 0.0088 & -0.0182 \\
 & SentiLARE & 0.0846 & -0.1719 \\
 & SentiCSE & 0.4786 & -1.0043 \\ 
 & \textbf{CARL} & \textbf{0.2649} & \textbf{-0.6063} \\
\hline
\end{tabular}%
}
\label{tab:alignment-uniformity}
\end{center}
\vspace{-3mm}
\end{table}

Our comprehensive ablation studies presented in Table~\ref{tab:sentiment-analysis} reveal that removing either the token perturbation detection mechanism or the momentum continuous contrastive learning objective led to notable performance degradation across all tasks. Specifically, in BERT, removing the token perturbation detection (w/o TP) resulted in a 2.5\% drop, while removing the momentum continuous contrastive learning (w/o MCCL) led to a 4.2\% decrease in average performance across the polarity classification task. 
These findings highlight the importance of both components in improving the quality of the learned representations and demonstrate their complementary roles.
\vspace{-5pt}

\begin{table}[H] 
\caption{Ablation study}
\label{tab:sentiment-analysis}
\vspace{-5pt}
\centering
\resizebox{\columnwidth}{!}{%
\begin{tabular}{|c|c|c|c|c|c|c|c|c|c|c|}
\hline
& \multicolumn{4}{c|}{\textbf{SentEval}} & \multicolumn{3}{c|}{\textbf{Valence}} & \multicolumn{3}{c|}{\textbf{Arousal}} \\
\cline{2-11}
Model & CR & MR & SST5 & IMDB & \textbf{$r$} & \textbf{$\rho$} & \textbf{MAE} & \textbf{$r$} & \textbf{$\rho$} & \textbf{MAE} \\
\hline
\textbf{CARL} &  \textbf{89.17} & \textbf{86.59} & \textbf{49.83} & \textbf{88.64} & \textbf{0.72} & \textbf{0.71} & \textbf{0.09} & \textbf{0.63} & \textbf{0.68} & \textbf{0.10} \\
w/o TP & 88.45 & 82.21  & 47.56 & 88.26 & 0.69 & 0.70 & 0.10 & 0.62 & 0.67 & 0.10 \\
w/o MCCL & 87.62 & 81.86 & 44.96 & 86.60 & 0.70 & 0.69 & 0.09 & 0.60 & 0.66 & 0.11 \\
Base RWA&86.06 & 79.46 & 43.78 & 86.66 & 0.67 & 0.66 & 0.10 & 0.59 & 0.64 & 0.11 \\
\hline 
\textbf{CARL} & \textbf{91.42} & \textbf{88.49} & \textbf{52.34} & \textbf{92.41} & \textbf{0.74} & \textbf{0.73} & \textbf{0.08} & \textbf{0.63} & \textbf{0.69} & \textbf{0.09} \\
w/o TP & 89.17 & 85.79 & 50.43 & 91.86 & 0.69 & 0.68 & 0.10 & 0.63 & 0.69 & 0.10 \\
w/o MCCL& 90.56 & 85.43  & 48.39 & 91.42 & 0.72 & 0.69 & 0.09 & 0.62 & 0.66 & 0.11 \\
Base RWA& 80.82 & 75.59 & 41.49 & 85.36 & 0.58 & 0.57 & 0.11 & 0.52 & 0.56 & 0.12 \\
\hline
\end{tabular}
}
\end{table}

\section{Conclusion}
In this work, we presented a novel framework for enhancing emotion-aware text embeddings by integrating momentum continuous label contrastive learning and gradient-based token perturbation detection. The experimental result demonstrates improved performance over baseline models in affective recognition tasks with well-aligned and uniformly distributed embeddings, validated across multiple benchmarks. This framework is useful for emotion-aware natural language processing, with potential applications spanning affective computing, text retrieval problems, and human-computer interaction.



\begin{thebibliography}{00}
\vspace{-5pt}
\bibitem{Afzal2023} S. Afzal, H. A. Khan, M. J. Piran, and J. W. Lee, A comprehensive survey on affective computing: Challenges, trends, applications, and future directions, IEEE Access, vol. 12, pp. 1–21, 2023.

\bibitem{Cambria2023} E. Cambria and A. Hussain, Sentiment analysis in the age of generative AI, IEEE Intelligent Systems, vol. 38, no. 1, pp. 12-16, 2023.

\bibitem{ip1} I. Park, D. Kim, and D. Har, MAC achieving low latency and energy efficiency in hierarchical M2M networks with clustered nodes, IEEE Sensors Journal, vol. 15, no. 3, pp. 1657-1661, 2015.

\bibitem{sur1} S. Kotte, J. R. K. Dabbakuti, Smart 6G sensor network based human emotion analysis by machine learning architectures, Wireless Pers Commun, 2024.

\bibitem{EmoRec} Qin, Xiangyu, et al., BERT-ERC: Fine-tuning BERT is enough for emotion recognition in conversation, Proceedings of the AAAI Conference on Artificial Intelligence, Vol. 37, No. 11, 2023.


\bibitem{KimReview} T. Kim, L. F. Vecchietti, K. Choi, S. Lee, and D. Har, Machine Learning for Advanced Wireless Sensor Networks: A Review, IEEE Sensors Journal, vol. 21, no. 11, pp. 12379-12397, 2021

\bibitem{BERT} J. Devlin., M. Chang, K. Lee, and K. Toutanova, BERT: Pre-training of deep bidirectional transformers for language understanding, Proceedings of the 2019 Conference of the North American Chapter of the Association for Computational Linguistics: Human Language Technologies, Volume 1, pp. 4171–4186, 2019.

\bibitem{GPT} A. Radford, K. Narasimhan, T. Salimans, and I. Sutskever, Improving language understanding by generative pre-training, OpenAI Technical Report, 2018.

\bibitem{pretraining} B. Li, H. Zhou, J. He, M. Wang, Y. Yang, and L. Li, On the sentence embeddings from pre-trained language models, the Conference on Empirical Methods in Natural Language Processing, 2020, pp. 9119–9130

\bibitem{BERT-PT} H. Xu, L. Zhang, J. Qian, and L. He, BERT post-training for review reading comprehension and aspect-based sentiment analysis, the Conference of the North American Chapter of the Association for Computational Linguistics, 2019.

\bibitem{SentiBERT} D. Yin, F. Xie, W. Zhang, and X. Li, SentiBERT: A transferable transformer-based architecture for compositional sentiment semantics, the 58th Annual Meeting of the Association for Computational Linguistics, 2020.

\bibitem{SentiLARE} P. Ke, T. Huang, and Y. Liu, SentiLARE: Sentiment-aware language representation learning with linguistic knowledge, the Conference on Empirical Methods in Natural Language Processing, 2020.

\bibitem{SentiX} Z. Zhou, L. Zhang, X. Chen, and J. Wang, SentiX: A sentiment-aware pre-trained model for cross-domain sentiment analysis, Proceedings of the 30th International Conference on Computational Linguistics, 2020.

\bibitem{Gao2021} T. Gao, X. Yao, and D. Chen, SimCSE: Simple contrastive learning of sentence embeddings, the Conference on Empirical Methods in Natural Language Processing, 2021.

\bibitem{Li2021} Z. Li, Y. Zou, C. Zhang, Q. Zhang, and Z. Wei. Learning implicit sentiment in aspect-based sentiment analysis with supervised contrastive pretraining. Proceedings of the Conference on Empirical Methods in Natural Language Processing, 2021.

\bibitem{SentiCSE} J. Kim, Y. Na, K. Kim, S. R. Lee, and D. K. Chae, SentiCSE: A sentiment-aware contrastive sentence embedding framework with sentiment-guided textual similarity, Proceedings of the 30th International Conference on Computational Linguistics, 2024.

\bibitem{grill2020bootstrap} Grill, Jean-Bastien, et al., Bootstrap your own latent — a new approach to self-supervised learning. Advances in neural information processing systems 33 : 21271-21284, 2020.

\bibitem{Russell2022} J. A. Russell, Revisiting the circumplex model of affect, Emotion Review, vol. 14, no. 1, pp. 3-10, 2022.

\bibitem{ELECTRA} K. Clark, M. Luong, Q.V. Le, and C.D. Manning, ELECTRA: Pre-training text encoders as discriminators rather than generators, International Conference on Learning Representations, 2020

\bibitem{PGD} A. Madry, A. Makelov, L. Schmidt, D. Tsipras, and A. Vladu, Towards deep learning models resistant to adversarial attacks, International Conference on Learning Representations, 2018.

\bibitem{EmoBank} Buechel, Sven, and Udo Hahn, Emobank: Studying the impact of annotation perspective and representation format on dimensional emotion analysis, Proceedings of the 15th Conference of the European Chapter of the Association for Computational Linguistics: Volume 2, 2017

\bibitem{IEMOCAP} C. Busso et al., IEMOCAP: interactive emotional dyadic motion capture database. Language Resources and Evaluation, 42, 335-359, 2018

\bibitem{FacebookPosts} D. Preotiuc-Pietro et al., Modelling Valence and Arousal in Facebook posts, Proceedings of the 7th workshop on computational approaches to subjectivity, sentiment and social media analysis, 2016

\bibitem{EmoTales} V. Francisco, R. Hervás, F. Peinado, P. Gervás. EmoTales: creating a corpus of folk tales with emotional annotations, Language Resources and Evaluation, 46, 341 - 381, 2011.

\bibitem{SentEval} A. Conneau and D. Kiela, SentEval: An evaluation toolkit for universal sentence representations, Language Resources and Evaluation, 2018.

\bibitem{GoEmotions} D. Demszky, E. Go, C. Shinn, and S. Kiritchenko, GoEmotions: A dataset of fine-grained emotions, the 58th Annual Meeting of the Association for Computational Linguistics, 2020.

\bibitem{wang2020understanding} T. Wang and P. Isola, Understanding contrastive representation learning through alignment and uniformity on the hypersphere, International Conference on Machine Learning, 2020.

\end{thebibliography}
\end{document}